\documentclass{svproc}
\usepackage{url}
\usepackage{graphicx}
\usepackage{graphics}
\usepackage{multicol}
\usepackage{footmisc}
\usepackage{amssymb}
\usepackage{amsmath}
\usepackage{algorithm,algpseudocode}

\begin{document}
\mainmatter              
\title{Simultaneous Skull Conductivity and Focal Source Imaging from {EEG} Recordings with the help of Bayesian Uncertainty Modelling}
\titlerunning{Skull Conductivity and Focal Source Imaging from {EEG} Recordings}  
%
\author{Alexandra Koulouri\inst{1} \and Ville Rimpil\"ainen\inst{2}}
\authorrunning{A. Koulouri and V. Rimpil\"ainen} 

\institute{Tampere University, Faculty of Information Technology and
Communication Sciences, Tampere, Finland,
\email{alexandra.koulouri@tuni.fi}\\ \and University of Bath,
Department of Physics, Bath, United Kingdom}

\maketitle              

\begin{abstract}
The electroencephalography (EEG) source imaging problem is very sensitive to the electrical
modelling of the skull of the patient under examination.
Unfortunately, the currently available EEG devices and their
embedded software do not take this into account; instead, it is
common to use a literature-based skull conductivity parameter.
In this paper, we propose a statistical method based on the Bayesian approximation error approach to compensate
for source imaging errors due to the unknown skull conductivity and,
simultaneously, to compute a low-order estimate for the actual skull
conductivity value. By using simulated EEG data that corresponds to
focal source activity, we demonstrate the potential of the method
to reconstruct the underlying focal sources and low-order errors
induced by the unknown skull conductivity. Subsequently, the estimated
errors are used to approximate the skull conductivity. The results
indicate clear improvements in the source localization accuracy and
feasible skull conductivity estimates.
\end{abstract}

\section{Introduction}
The estimation of the focal brain activity from
electroencephalography (EEG) data is an ill-posed inverse problem,
and the solution depends strongly on the accuracy of the head model, particularly the used skull conductivity value \cite{Vanrumste2000,dan11}.
The skull conductivity value can be calibrated 
by using well defined somatosensory evoked potentials / fields in combination with EEG \cite{lew09b}, combined EEG/MEG \cite{ayd14}
or by using electrical impedance tomography (EIT) -based techniques \cite{gon03a}.
However, these methods are sometimes unreliable, computationally effortful
and may require auxiliary imaging tools.

In our previous studies \cite{rim17,Rimpilaeinen2019}, we demonstrated
a more straightforward approach that was based on 
uncertainty modelling via the Bayesian approximation error (BAE) approach \cite{kaipio2013}.
With BAE, the natural variability of the skull conductivity was
modelled in a probability distribution, and subsequently its effects
on the EEG data were taken statistically into account in the
observation model (likelihood) with the help of an additive
modelling error term. This allowed us to use an approximate skull
conductivity in the EEG source imaging, i.e. to neglect the
calibration step completely, and still get feasible source configuration estimates.

In this paper, we demonstrate with simulations that in addition to the source
configuration the Bayesian uncertainty modelling also allows simultaneous
estimation of the unknown skull conductivity. This estimation can be
carried out through a low order decomposition of the modelling
error term \cite{kaipio2013,Nissinen2011a}. 

We envision that the proposed approach could be used as such, or alternatively as a
calibration step to determine the skull conductivity of a patient. The benefit compared to 
current injection -based techniques is that it relies solely on EEG data and
thus avoids the requirement for a brain rest period before the
actual EEG examination. In addition, this approach does not add computational
effort (to the regular source imaging) since it does not involve non-linear formulations or
unstable matrix inversions \cite{lew09b,Papageorgakis2017}; instead, all the required pre-computations 
can be performed off-line before any experiments take place.

\section{Theory}

\subsection{Bayesian framework with linear forward model}

The computational domain is denoted with $\Omega$ and its electric conductivity with $\sigma(x)$ where $x \in \Omega$. The numerical
observation model is
\begin{equation}
\label{eq:obm2}
v = A(\sigma) d + e,
\end{equation}
where $v \in \mathbb{R}^m$ are the measurements, $m$ is the number
of measurements, $A(\sigma) \in \mathbb{R}^{m \times 2n}$ is the
(2-dimensional) lead field matrix that depends on $\sigma$, $d\in
\mathbb{R}^{2n}$ is the distributed dipole source configuration and
$e \sim \mathcal{N}(e_*,\Gamma_e)$ is the measurement noise. Note
that here the model $A(\sigma)$ assumes (unrealistically) that the
accurate values of electric conductivities are known.

In the Bayesian framework, the inverse
solution is the posterior density
\begin{equation}\label{post}
\pi(d|v) \propto \pi(v|d)\pi(d),
\end{equation}
where $\pi(v|d)$ is the likelihood and $\pi(d)$ the prior.

For the model (\ref{eq:obm2}), the likelihood can be
written as
\begin{equation}\label{lik1}
\pi(v|d) \propto
\exp\Big(-\frac{1}{2}(v-A(\sigma)d-e_{*})^\mathrm{T}\Gamma_e^{-1}(v-A(\sigma)d-e_{*})\Big).
\end{equation}

\subsection{Bayesian approximation error (BAE) approach}

In BAE, the accurate lead field matrix
$A(\sigma)$ is replaced with an approximate (standard) lead field
matrix $A_{0} = A(\sigma_0)$ in which fixed electric conductivities
$\sigma_0$ are used. Now, the observation model (\ref{eq:obm2}) can
be written as
\begin{equation}\label{eq:AMEA}
v= A_{0}d+\varepsilon+e
\end{equation}
where 
\begin{equation}\label{eq:approximationerror}
\varepsilon = A(\sigma) d-A_0d \end{equation} is the induced {\em
approximation error}, $\varepsilon \in \mathbb{R}^m$.

In order to approximate the unknown skull conductivity,  we need to
obtain an estimate about the approximation error. 
Similarly as in \cite{Nissinen2011a,kaipio2013}, we express
$\varepsilon$ as a linear combination of basis functions.
Numerically, the estimation of $\varepsilon$ is tractable if
most of its variance  is captured in only few of the decomposition
terms \cite{Nissinen2011a,kaipio2013}.

A feasible set of basis functions that assigns most of the variance of $\varepsilon$
in the first few terms can be chosen based on the eigenvalue decomposing of the modelling error
covariance $\Gamma_{\varepsilon} =\mathbb{E}
[(\varepsilon-\varepsilon_*)(\varepsilon-\varepsilon_*)^{\mathrm{T}}]$.
 In particular, $\Gamma_{\varepsilon}$ can be decomposed according
 to
\begin{equation}
\Gamma_{\varepsilon} = \Sigma_{k=1}^{m} \lambda_k w_k
w_k^{\mathrm{T}},
\end{equation}
where $\lambda_k\in \mathbb{R}^m$ are the eigenvalues and $w_k \in
\mathbb{R}^m$ are the eigenvectors of $\Gamma_\varepsilon$. Based on
this decomposition, we have that
$\varepsilon-\varepsilon_*\;\in\;\mathrm{span}\{w_1,\ldots,w_k\}$.

The approximation error can be expressed as
\begin{equation}\varepsilon = \varepsilon_* + \varepsilon' +
\varepsilon'',\end{equation} where $\varepsilon'=\Sigma_{k=1}^{p}
\alpha_k w_k$ and $\varepsilon''=\Sigma_{j=p+1}^{m} \beta_j w_j$,
with $\mathbb{E}[\varepsilon'] = 0$ and $\mathbb{E}[\varepsilon''] =
0$. The coefficients $\alpha_k$ and $ \beta_j$ are given by the
inner products $\alpha_k=<\varepsilon - \varepsilon_*,w_k>$ and
$\beta_j=<\varepsilon - \varepsilon_*,w_j>$.

Thus, the observation model (\ref{eq:AMEA}) can be rewritten as
\begin{equation}
\label{eqom}
v = A_0 d + \varepsilon_* + W \alpha + \varepsilon'' + e,
\end{equation}
where $\varepsilon'=W \alpha $, $W = [w_1,w_2,...,w_p] \in
\mathbb{R}^{m \times p}$ contains the first $p$ eigenvectors and
$\alpha = (\alpha_1, \alpha_2,...,\alpha_p)^{\mathrm{T}} \in
\mathbb{R}^p$. Note that $\varepsilon'' = Q \beta =
QQ^\mathrm{T}\varepsilon$ where $Q = [w_{p+1},\ldots,w_m]\in
\mathbb{R}^{(m-p)\times m}$.

For the simultaneous estimation of $d$ and $\alpha$, we have to
construct the posterior model $\pi(d,\alpha|v)$. To obtain a
computationally efficient solution, we make the technical
approximation that $(d,\alpha,e,\varepsilon'')$ are mutually
Gaussian and uncorrelated \cite{Nissinen2011a,kaipio2013}.
Then, we obtain the approximate likelihood
\begin{eqnarray}\label{lik2}
&&\tilde{\pi}(v | d, \alpha) \propto \exp
\Big(-\frac{1}{2}(v-A_{0}d-\varepsilon_{*}-W\alpha-e_{*})^\mathrm{T} \nonumber \\
&& \qquad
(\Gamma_{\varepsilon''}+\Gamma_e)^{-1}(v-A_{0}d-\varepsilon_{*}-W\alpha-e_{*})\Big),
\end{eqnarray}
where $\Gamma_{\varepsilon''} = \Sigma_{p+1}^{m}\lambda_j w_j
w_j^{\mathrm{T}}$. Based on the eigenvalue decomposition and the
Gaussian approximations, we have that the coefficients $\alpha$ are
normally distributed with zero mean and covariance $\Gamma_{\alpha}=
\mathrm{diag}(\lambda_1,\ldots,\lambda_p)$ (where
$\lambda_1>\ldots>\lambda_p$). Thus, the posterior density becomes
\begin{equation}\label{eq:posteriorBAE}
\tilde{\pi}(d,\alpha|v) \propto \mathrm{exp}
\left(-\frac{1}2L_{\varepsilon''+e}\|v-A_0d-W\alpha-\varepsilon_*-e_*\|_2^2\right)
\mathrm{exp}\left(-\frac{1}2L_\alpha\|\alpha\|_2^2\right)\;\pi(d),
\end{equation}
where the Cholesky factors are
$(\Gamma_{\varepsilon''}+\Gamma_{e})^{-1}=L_{\varepsilon''+e}^\mathrm{T}
L_{\varepsilon''+e}$ and $\Gamma_\alpha^{-1} = L_
\alpha^{\mathrm{T}}L_\alpha$.


\section{Methods}

\subsection{Head models}
\label{hm}
\begin{figure}[ht]
      \centering
          \includegraphics[width=0.99\columnwidth]{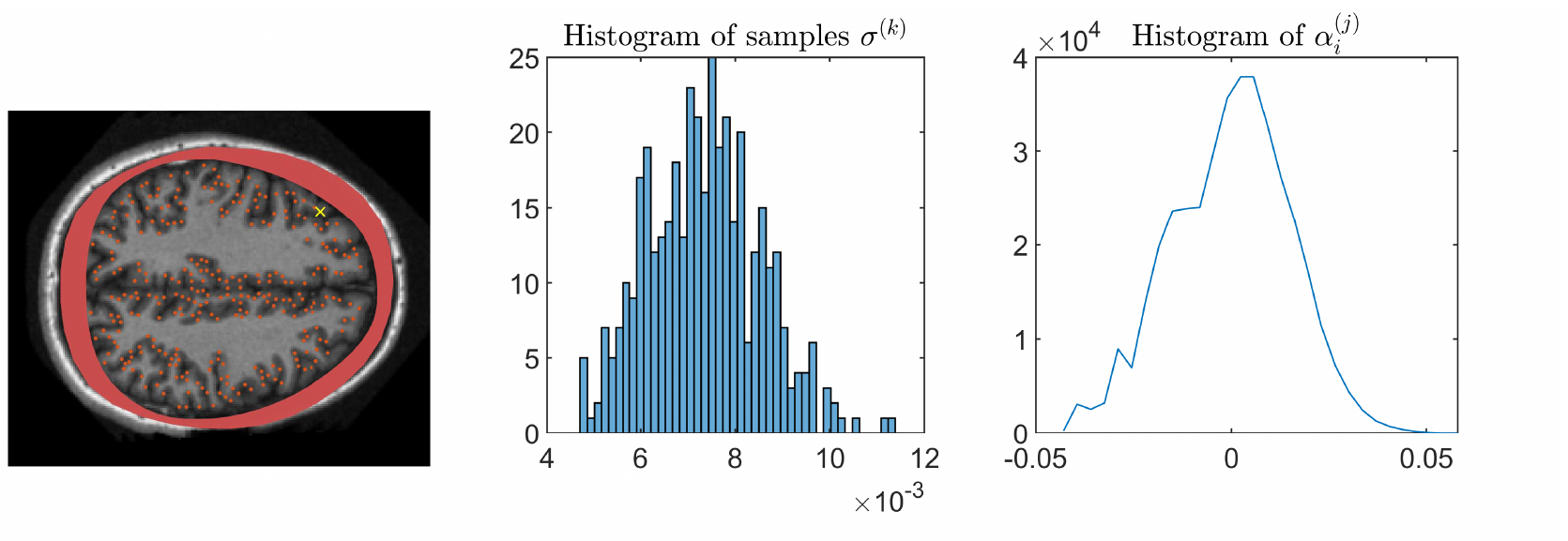}
      \caption{Left: The 2-dimensional human head model used in the computations was segmented in scalp, skull (illustrate with solid red color) and brain compartments. The source space locations are illustrated with red dots. Middle: Histogram of 400 skull conductivity samples. Right: To sample $\alpha$, first a source location (shown with a yellow cross on the left) and 1000 amplitude values for the source are drawn, then sample lead fields (based on the 400 skull conductivities) are used to compute the corresponding approximation errors. The resulting histogram of the first $\alpha$ parameter ($k=1$) has approximately Gaussian form.}
      \label{figure1setup}
\end{figure}

A 2-dimensional finite element (FE) -based three compartment (scalp, skull and brain) head model
with 32 electrodes was used in the demonstrations (see, Figure \ref{figure1setup}). For the forward
simulations and inverse estimations two different meshes with 2518 and
1780 nodes were constructed, respectively, with source spaces covering
the gray matter of the brain.

In particular, we created $K=400$ head models with skull
conductivity samples $\sigma^{(k)}$ drawn from a Gaussian
distribution $\pi(\sigma)$ with mean $\sigma_*= 0.0073$ (in S/m) and
standard deviation 0.0013 (see, Figure \ref{figure1setup}). We refer to these head models as {\it
sample} head models. This skull conductivity distribution coincides
with the literature values that range between 0.0041 and 0.033
\cite{hoe03,hom95,ayd14}. The electric conductivity of the scalp was
0.43 \cite{ram06} and the brain 0.33 \cite{ram06}. We also created a
\textit{standard} head model with skull conductivity
$\sigma_0=0.0085$.

\subsection{Computation of the approximation error
statistics}\label{sec:sampling}
\label{ces} 

The approximation error (\ref{eq:approximationerror}) depends both
on the skull conductivity and the dipole source location. 
Feasible estimates for the skull conductivity can be expected if
most of the variance of $\varepsilon$ is related to the skull
uncertainty. Therefore, we isolated the effect that the skull  has on
the observations. To do that, we estimated the error statistics for each dipole
location separately. In the following text, the subscript $i$  denotes a
specific dipole location.

The samples for the error statistics at location $i$ were created by
first choosing randomly one of the \emph{sample} head models and
evaluating both the sample model and the standard model with the
same single dipole source $d_i^{(s)}$, i.e.
\begin{equation}
\varepsilon^{(j)}_i = A(\sigma^{(k)}){d}_i^{(s)} - A_{0}d_i^{(s)}.
\end{equation}

A set of $S=1000$ single radial dipole samples $d_i^{(s)}$ with
varying amplitudes were drawn from a Gaussian distribution $\pi(d_i)$
with mean 1 and standard deviation 0.01,
and 200 different leadfield matrices were used to estimate the error statistics at location $i$. 
Superscript $j =s+S(k-1)$ where $s=1,\ldots,S$ and $k=1,\ldots K$.
The mean $\varepsilon_{i*}$ and error covariance
$\Gamma_{\varepsilon_i}$ at location $i$ were
\begin{equation}\label{eq:MeanAndCovariance}
\varepsilon_{i*} = \frac{\sum_{j=1}^{J}{\varepsilon}_i^{(j)}}{J}
\,\,\mbox{and}\,\,\Gamma_{\varepsilon_i}=
\frac{1}{J-1}\sum_{j=1}^{J}(\varepsilon_i^{(j)}-\varepsilon_{i*})(\varepsilon_i^{(j)}-\varepsilon_{i*})^\mathrm{T}.
\end{equation}
Samples $\alpha_i^{(j)}$ and $\varepsilon_i''^{(j)}$ were evaluated
according to
$\alpha_i^{(j)} =
W_i^\mathrm{T}(\varepsilon_i^{(j)}-\varepsilon_{i*})$ and
$\varepsilon_i''^{(j)} =
Q_iQ_i^{\mathrm{T}}(\varepsilon_i^{(j)}-\varepsilon_{i*})$, where
$W_i$ and $Q_i$ were matrices with columns the first $p$ and
remaining eigenvectors of $\Gamma_{\varepsilon_i}$ respectively.

\subsection{Simultaneous skull conductivity and source estimation}

In this work,
we compute the maximum a posteriori (MAP) estimates of the posterior
(\ref{eq:posteriorBAE}). Since we study only single dipole source
cases, we can use a modified single dipole scan algorithm and solve
\begin{equation}
\begin{bmatrix}
\hat{d}_i\\ \hat{\alpha}_i
\end{bmatrix}:=\begin{bmatrix}L_{\varepsilon_i''+e} A_0^i & L_{\varepsilon_i''+e}W_i\\ \mathbf{0} & L_{\alpha_i}\end{bmatrix}^{-1} \begin{bmatrix}L_{\varepsilon_i''+e} (v -\varepsilon_{i*}-e_*)\\
0\end{bmatrix},\;\;\;\mbox{for } i=1,\ldots,n,
\end{equation}
where $n$ is the total number of source locations, $A_0^i=[A_{0x}^i
A_{0y}^i ]$ are the columns of the (standard) leadfield matrix corresponding to
location $i$, $\hat{d}_i=(d_{x_i},d_{y_i})$ is the dipole reconstructed at
location $i$, $W_i\in\mathbb{R}^{m\times p}$ are the basis functions
obtained from the eigenvalue decomposition of the error covariance
$\Gamma_{\varepsilon_i}\in\mathbb{R}^{m\times m}$, $\alpha_i\in
\mathbb{R}^p$ are the reconstructed coefficients for these basis functions, and
$L_{\alpha_i}=\frac{1}2\mathrm{diag}(\lambda_1^{-1},\lambda_2^{-1},...,\lambda_p^{-1})$
where $\lambda$ parameters are the eigenvalues of the covariance
$\Gamma_{\varepsilon_i}$ (\ref{eq:MeanAndCovariance}). The number of
$\alpha$ parameters, as well as, the number of eigenvalues (used in
the estimates) was defined based on the minimum value $p$ that
satisfied $\frac{\sum_{k=1}^p\lambda_k}{\sum_{k=1}^m\lambda_k}\geq
0.85$. In most cases, either one or two $\lambda$ parameters were needed.

The solution is the pair 
\begin{equation}\label{eq:Functional3}
(\hat{d}_l,\hat{\alpha}_l)\leftarrow\min_{i=1:n}
\{\|L_{\varepsilon_i''+e}(v-A_0^i
\hat{d}_i-\varepsilon_{i*}-W_i\hat{\alpha}_i-e_{*})\|_2^2+L_{\alpha_i}\|\hat{\alpha}_i\|_2^2
\}
\end{equation}
where $i=l$ is the location that gives the lowest value for the functional
(\ref{eq:Functional3}).

Given the dipole location $l$, the approximation error
(\ref{eq:approximationerror}) can be written as
\begin{equation}\label{eq:approximationError_l} \varepsilon_l=A^l(\sigma)d_l-A_0^l
d_l,\end{equation} where
$\varepsilon_l=W_l\alpha_l$. 

The skull conductivity can now be estimated by maximizing
\begin{equation} \label{eq:MAPsigma}
\hat{\sigma}:=\max_{\sigma}\pi(\sigma|\alpha_l).\end{equation} The
conditional probability distribution $\pi(\sigma|\alpha_l)$ is
\begin{equation}
\pi(\sigma|\alpha_l) = \int \pi(\sigma,d_l|\alpha_l) \;\mathrm{d}d_l
\propto \pi(\sigma) \int \pi(\alpha_l|\sigma,d_l) \pi(d_l) \;
\mathrm{d}d_l,
\end{equation}
where $\sigma$ and $d_l$ are mutually independent.

From equation (\ref{eq:approximationError_l}), we have that $\int
\pi(\alpha_l|\sigma,d_l) \pi(d_l) \; \mathrm{d}d_l=\int \delta(W_l\alpha_l-[A^l(\sigma)-A_0^l]d_l) \pi(d_l) \; \mathrm{d}d_l
=\frac{1}{|f_l^{-1}(\sigma)|}\pi_{d_l}(f_l(\sigma)\alpha_l)$ where the subscript $d_l$ is used to clarify 
that this is the probability density of $d_l$, and the expression $f_l(\sigma)$ formally
comes from the inversion of the lead field matrix difference and
multiplication with $W_l$.
So, $\pi(\sigma|\alpha_l)\propto \pi(\sigma)\;
\frac{1}{|f_l^{-1}(\sigma)|}\pi_{d_l}(f_l(\sigma)\alpha_l)$. Even though $\pi(\sigma)$ an
$\pi_{d_l}(.)$ have been modelled as Gaussian distributions, the
non-linearity with respect to $\sigma$ and often the infeasibility
of estimating $f_l(\sigma)$ can prohibit a straightforward $\sigma$
estimation. Instead we can use
$\pi(\sigma|\alpha_l)\propto\pi(\sigma,\alpha_l)$.

Since, $\pi(\sigma,\alpha_l)=\frac{1}{|f_l^{-1}(\sigma)|}\pi(\sigma)\;
\pi_{d_l}(f_l(\sigma)\alpha_l)$,
 the joint distribution of $\sigma$
and $\alpha_l$ can be approximated as Gaussian, and therefore the
MAP estimate (\ref{eq:MAPsigma}) gives
\begin{equation}
\hat{\sigma} = \sigma_* + \Gamma_{\sigma \alpha_l}
\Gamma_{\alpha_l}^{-1} \hat{\alpha}_l,
\end{equation}
where $\hat{\alpha}_l$ is the solution of problem
(\ref{eq:Functional3}), $\sigma_*$ is the mean value of the
postulated skull conductivity distribution,
 $\Gamma_{\sigma \alpha_l}$ is the cross-covariance between $\sigma$ and $\alpha_l$
 (estimated using samples $\alpha_l^{(j)}$ and $\sigma^{(k)}$ as described in section~\ref{sec:sampling}),
 and $\Gamma_{\alpha_l}=\mathrm{diag}(\lambda_1,\ldots,\lambda_p)$ where $\lambda_p$ are the first $p$ eigenvalues of the sampled-based error covariance $\Gamma_{\varepsilon_l}$ (\ref{eq:MeanAndCovariance}).

\section{Results, discussion and  future work}

\begin{figure}[hp]
      \centering
          \includegraphics[width=0.80\columnwidth]{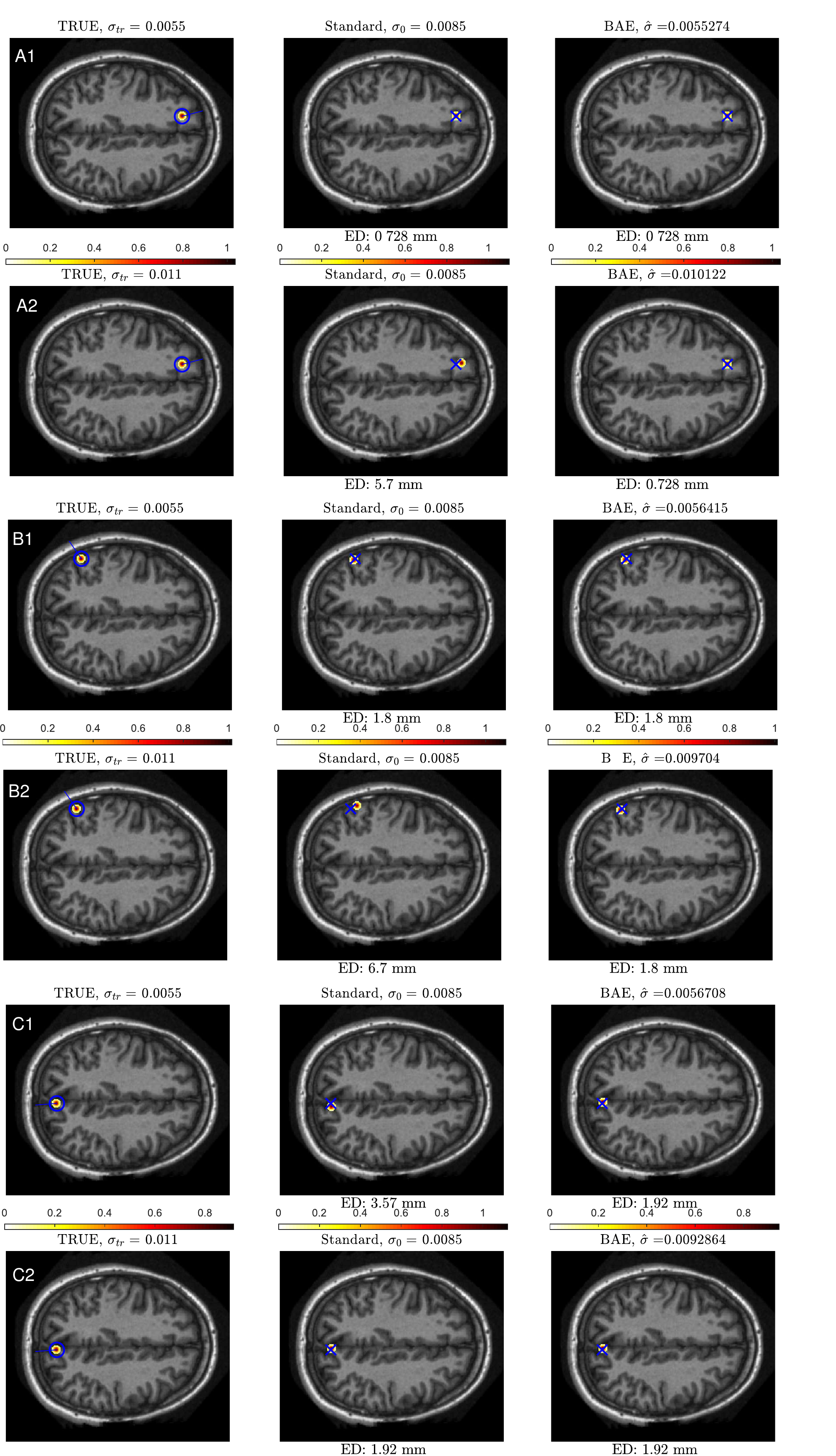}
      \caption{Left column: Test cases with three different source locations and two different skull conductivity values, either lower (0.0055 S/m) or higher (0.011 S/m) than the standard skull conductivity (0.0085 S/m). The blue circle and the line represent the true location and direction of the dipole source, respectively. Middle column: Source reconstruction results when the standard skull conductivity was used in the lead field. The ED values give the Euclidean distance between the true and reconstructed source. The true location is denoted with a blue cross. Right column: Simultaneous skull conductivity and source location reconstructions with the proposed Bayesian uncertainty modelling approch. As can be seen, the proposed approach improves the source localization compared to the results with the standard model and additionally achieves to  estimate the skull conductivity.}
      \label{figure1}
\end{figure}

To demonstrate the simultaneous dipole and skull conductivity
reconstruction, we used two test cases with different skull
conductivities. The EEG data was computed using the {\it accurate}
model that had either skull conductivity 0.0055 or 0.011 S/m, a
single radial dipole source, and signal to noise
ratio 30 dB. In Figure \ref{figure1}, we show first the test set-up,
then the dipole-scan reconstruction with the standard lead field that has skull conductivity 
0.0085 S/m, and
finally the proposed Bayesian error modelling result. To ease comparisons, 
we calculated the Euclidean distance (ED) (in milli meters)
between the actual and reconstructed source. It can be seen that the 
proposed uncertainty modelling improves the source
reconstructions when compared to the solution of the standard model.
Moreover, the estimates for the skull conductivity are close to
the true skull conductivities.

The feasibility of the skull conductivity estimates
depends on how well the first few eigenvectors of
$\Gamma_{\varepsilon}$ describe the conductivity related modelling
errors. In the future, we aim to isolate the conductivity related errors
from other error sources (e.g. discretization) in order to improve the skull conductivity estimates.

In our analysis the head geometry was considered fully known;
however, it can also be taken into account as an uncertainty
\cite{kou16}. We also
note that in this work we estimated a single conductivity value for
the whole skull. As  a next step, we plan to use the proposed approach
in conjugation with other optimization techniques
\cite{Acar2016,Costa2016} to produce more detailed skull conductivity
maps.

\section{Conclusion}

We have demonstrated that with the help of the Bayesian uncertainty
modeling the EEG source imaging problem could be reformulated in such a
way that it allows to estimate, simultaneously with the source
configuration, a low-order approximation for the modelling error.
With the help of this modelling error estimate, the underlying skull
conductivity could be reconstructed. 

\section*{Acknowledgements}

This project has received funding from the ATTRACT project funded by the EC under Grant Agreement 777222 and from the Academy of Finland post-doctoral program (project no. 316542).

\section*{Conflict of interest}

The authors declare that they have no conflicts of interest.

\bibliographystyle{spphys}

\begin{thebibliography}{10}
\providecommand{\url}[1]{{#1}}
\providecommand{\urlprefix}{URL }
\expandafter\ifx\csname urlstyle\endcsname\relax
  \providecommand{\doi}[1]{DOI \discretionary{}{}{}#1}\else
  \providecommand{\doi}{DOI \discretionary{}{}{}\begingroup
  \urlstyle{rm}\Url}\fi

\bibitem{Vanrumste2000}
B.~Vanrumste, G.V. Hoey, R.V. de~Walle, M.~D'Have, I.~Lemahieu, P.~Boon, Med.
  Biol. Eng. Comput. \textbf{38}(5), 528 (2000).

\bibitem{dan11}
M.~Dannhauer, B.~Lanfer, C.~Wolters, T.~Kn\"osche, Hum. Brain Mapp.
  \textbf{32}, 1383 (2011)

\bibitem{lew09b}
S.~Lew, C.H. Wolters, A.~Anwander, S.~Makeig, R.S. MacLeod, Hum. Brain Mapp.
  \textbf{30}(9), 2862 (2009).

\bibitem{ayd14}
U.~Aydin, J.~Vorwerk, P.~K\"upper, M.~Heers, H.~Kugel, A.~Galka, L.~Hamid,
  J.~Wellmer, C.~Kellinghaus, S.~Rampp, C.H. Wolters, PLoS ONE \textbf{9},
  e93154 (2014)

\bibitem{gon03a}
S.I. Goncalves, J.C. de~Munck, J.P. Verbunt, R.M. Heethaar, F.H. da~Silva, IEEE
  Trans. Biomed. Eng. \textbf{50}(9), 1124 (2003)

\bibitem{rim17}
V.~Rimpil\"ainen, A.~Koulouri, F.~Lucka, J.P. Kaipio, C.H. Wolters, IFMBE Proc.
  \textbf{65}, 892 (2018)

\bibitem{Rimpilaeinen2019}
V.~Rimpil\"ainen, A.~Koulouri, F.~Lucka, J.P. Kaipio, C.H. Wolters, {NeuroImage}
  \textbf{188}, 252 (2019).

\bibitem{kaipio2013}
J.~Kaipio, V.~Kolehmainen, in \emph{Bayesian Theory and Applications}, ed. by
  P.~Damien, N.~Polson, D.~Stephens (Oxford University Press, 2013)

\bibitem{Nissinen2011a}
A.~Nissinen, V.~Kolehmainen, J.P. Kaipio, Int. J. Uncertain.
  Quantif. \textbf{1}(3), 203 (2011)

\bibitem{Papageorgakis2017}
C.~Papageorgakis, Patient specific conductivity models: characterization of the
  skull bones.
\newblock Ph.D. thesis (2017)

\bibitem{hoe03}
R.~Hoekema, G.H. Wieneke, C.W. {van Veelen}, P.C. {van Rijen}, G.J. Huiskamp,
  J.~Ansems, A.C. {van Huffelen}, Brain Topogr. \textbf{16}(1), 29 (2003)

\bibitem{hom95}
S.~Homma, T.~Musha, Y.~Nakajima, Y.~Okomoto, S.~Blom, R.~Flink, K.E. Hagbarth,
  Neurosci. Res. \textbf{22}(1), 51 (1995)

\bibitem{ram06}
C.~Ramon, P.H. Schimpf, J.~Haueisen, Biomed. Eng. Online \textbf{5}(10) (2006)

\bibitem{kou16}
A.~Koulouri, V.~Rimpil\"ainen, M.~Brookes, J.P. Kaipio, Appl. Num. Math.
  \textbf{106}, 24 (2016)

\bibitem{Acar2016}
Z.A. Acar, C.E. Acar, S.~Makeig, {NeuroImage} \textbf{124}, 168 (2016).

\bibitem{Costa2016}
F.~Costa, H.~Batatia, T.~Oberlin, J.Y. Tourneret, arXiv: 1609.06874  (2016)

\end{thebibliography}

\end{document}